\documentclass[accepted]{article}

\usepackage{times}
\usepackage{graphicx} 
\usepackage{subfigure} 

\usepackage{natbib}

\usepackage{algorithm}
\usepackage{algorithmic}

\usepackage{hyperref}

\usepackage{icml2015} 

\usepackage{pgfplots}
\pgfplotsset{compat=newest}

\usepackage{color}
\usepackage{balance}

\icmltitlerunning{Accuracy and fairness in binary classification}

\begin{document} 

\twocolumn[
\icmltitle{On the relation between accuracy and fairness in binary classification}

\icmlauthor{Indr\.e \v{Z}liobait\.e}{indre.zliobaite@aalto.fi}
\icmladdress{Aalto University and Helsinki Institute for Information Technology HIIT, Finland}


\vskip 0.3in
]

\begin{abstract} 
Our study revisits the problem of accuracy-fairness tradeoff in binary classification. 
We argue that  comparison of non-discriminatory classifiers needs to account for different rates of positive predictions, otherwise conclusions about performance may be misleading, because accuracy and discrimination of naive baselines on the same dataset vary with different rates of positive predictions. 
We provide methodological recommendations for sound comparison of non-discriminatory classifiers, and present a brief theoretical and empirical analysis of tradeoffs between accuracy and non-discrimination. 
\end{abstract} 

\section{Introduction}

Discrimination-aware machine learning is an emerging research area, which studies how to make predictive models free from discrimination, when historical data, on which they are built, may be biased, incomplete, or even contain past discriminatory decisions. Research assumes that the protected grounds, against which discrimination is forbidden, are given by legislation. The goal for machine learning is to develop algorithmic techniques for incorporating those non-discriminatory constraints into predictive models. 

A number of studies in discrimination-aware machine learning and data mining \cite{Pedrechi09,Kamiran10,Calders10} focus on achieving equal acceptance rates (proportions of positive decisions) for favored and protected groups of individuals in binary classification. Forcing acceptance rates to be equal without taking into account other characteristics of individuals can be seen as an affirmative action, which introduces positive discrimination promoting the protected community. This may be desired for legal and political reasons. 

We revisit this popular scenario of discrimination aware machine learning, and identify some pitfalls to avoid when comparing the performance of such classifiers, that is, a comparison may be misleading if the proportions of positive predictions of the classifiers are different. 
We provide methodological recommendations for sound comparison, and present a brief theoretical and empirical analysis of tradeoffs between accuracy and non-discrimination.

\section{Problem setting and assumptions}

Given a dataset that contains discrimination the goal is to build a classifier that would be as accurate as possible, and obey non-discrimination constraints. 
For example, a model could decide upon granting a loan given demographic information and financial standing, and considering ethnicity of an applicant (native, foreign) as the protected ground. We assume that the values of the target variable (labels) in the historical dataset are objectively correct, e.g. whether the loan has been repaid or not. 
For discrimination to happen the target variable needs to be polar, that is, one outcome (accept) should be preferred over the other (reject).

Let $X$ denote a set of input variables (e.g. salary, assets), $s$ denote the protected characteristic (e.g. ethnicity: native ($w$) or foreign ($b$)), and $y$ denote the target variable (e.g. loan decision: accept ($+$) or reject ($-$)). A classifier maps $X$ to $y$, that is, $\hat{y}  = f(X)$. Even though $s$ is not among the input variables, some variables in $X$ may be correlated with $s$ (e.g. social security payment history may be shorter for foreigners, because they have arrived recently), and, as a result, classifier $f$ may capture the protected characteristics, and induce indirect discrimination in decision making. 

Let  discrimination be measured as the difference in rates of acceptance: $d = p(+|w) - p(+|b)$.
Suppose that discrimination in the historical dataset is $d_0 = \delta$, the desired discrimination in the classifier output is $d^\star$, the proportion of favored individuals in the data is $p(w) = \alpha$, the prior probability of acceptance in the data is $p(+) = \pi_0$, and the rate of acceptance in the classifier output is $p_f(+) = \pi$.

Many classifiers produce probability scores (such as Naive Bayes or logistic regression). 
Typically, a probability score can be computed for non-probabilistic classifiers as well (such as kNN, SVM, decision trees).
Individuals scoring above a threshold, which by default is typically $0.5$, will get a positive decision.
Considering available resources a decision maker can choose a different threshold. 
Suppose that the objective is to keep discrimination at the desired level $d^\star$ (typically zero), and at the same time maximize the prediction accuracy. Effectively, by choosing the threshold a decision maker chooses the acceptance rate $\pi$.

%


\section{Accuracy and fairness}

The performance of discrimination-aware classifiers is typically compared by plotting discrimination vs. accuracy. 
 An attempt to remove discrimination can easily produce classifiers with different acceptance rates $\pi$ from those in the original dataset, especially when using off-the-shelve classifier implementations (e.g. WEKA\footnote{\url{http://www.cs.waikato.ac.nz/ml/weka/}}), which simply round the numerical probability scores without any constraints on the positive output rates.

\textbf{Our main message is that evaluation of non-discriminatory classifiers must take into account rates of acceptance, otherwise classifier performance is not comparable, because changing the acceptance rate changes baseline accuracy and baseline discrimination.} 

A small experiment with a benchmark dataset (Adult from UCI\footnote{\url{http://archive.ics.uci.edu/ml/}} repository) illustrates the situation. The target variable describes whether a person has high income or low. The protected characteristic (gender) is not among the inputs. We randomly split the dataset into two halves: training and testing. We train a logistic regression (similar results have been obtained with Naive Bayes and decision tree J48) on a train set, output class probability scores for the test set, and vary the classification threshold from $0$ to $1$, which changes the acceptance rate $\pi$.
We also plot the accuracy of a random classifier that does not use any inputs, but randomly decides upon the outcome given the probability of acceptance $\pi$.
 Figure \ref{fig:output} presents the results.
\definecolor{mycol1}{HTML}{0072BD}
\definecolor{mycol2}{HTML}{D95319}
\definecolor{mycol3}{HTML}{EDB120}
\definecolor{mycol4}{HTML}{7E2F8E}
\definecolor{mycol5}{HTML}{77AC30}
\definecolor{mycol6}{HTML}{4DBEEE}
\definecolor{mycol7}{HTML}{A2142F}
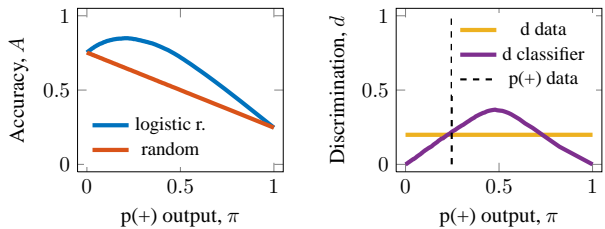
\begin{figure}
\centering
\begin{tikzpicture}[scale=0.8]
\begin{axis}[name = plot1,
xmin= -0.05, xmax=1.05, ymin=-0.05, ymax=1.05,
legend style={draw=none,font=\footnotesize},
legend pos= south west,
height=4.3cm, width=5cm, 
xlabel = {p(+) output, $\pi$}, ylabel = {Accuracy, $A$}]
\addplot[mycol1,line width = 2pt]  table[x = ppred0, y = acc0] {results_logistic.dat};
\addplot[mycol2,line width = 2pt]  table[x = ppred0, y = accr0] {results_logistic.dat};
\legend{logistic r., random};
\end{axis}
\begin{axis}[name=plot2, at=(plot1.right of south east), anchor=left of south west,
height=4.3cm, width=5cm, 
xshift=0.6cm,
xmin= -0.05, xmax=1.05, ymin=-0.05, ymax=1.05,
legend style={draw=none,font=\footnotesize},
legend pos= north east,
xlabel = {p(+) output, $\pi$}, ylabel = {Discrimination, $d$}]
\addplot[mycol3,line width = 2pt]  table[x = ppred0, y = ddif_data] {results_logistic.dat};
\addplot[mycol4,line width = 2pt]  table[x = ppred0, y = ddif0] {results_logistic.dat};
\addplot[black,dashed,line width = 1pt]  table[x = ptrue, y = nn] {results_logistic.dat};
\legend{d data, d classifier, p(+) data};
\end{axis}
\end{tikzpicture} 
\caption{Accuracy and discrimination measured directly.}
\label{fig:output}
\end{figure}

From the left plot we see that the more extreme the acceptance rate is (either all reject, or all accept), the closer the performance of an intelligent classifier (logistic regression) is to that of a random classifier, which assigns labels at random. Therefore, better observed accuracy does not necessarily mean better classification ability, if the acceptance rates of the two classifiers are different. In order to be able to compare such classifiers we could normalize the accuracy with respect to $\pi$. Therefore, we suggest using for comparison a normalized accuracy, such as Cohen's Kappa \cite{Cohen60}, which indicates by how much a classifier in question is better than a random classifier: 
\begin{equation}
\kappa = \frac{A - R}{1 - R},
\end{equation}
where $A$ is the accuracy of the classifier in question, and $R$ is the accuracy of a random classifier, in our case\\ $R = \pi_0\pi + (1-\pi_0)(1-\pi)$. Note, that $\kappa \in [0,1]$, where $1$ means the ideal accuracy, and $0$ indicates a random result\footnote{One could consider other accuracy measures for imbalanced data, such as F-score. We prefer Cohen's Kappa, since F-score does not behave consistently at the extreme acceptance rates, and, therefore, is more difficult to interpret. F-score of a classifier that accepts everybody would be equal to $\pi_0$, which varies depending on the dataset, while Kappa always gives $1$ in this case. }. 

In the right plot we see how discrimination varies with different acceptance rates. There is no discrimination if everybody is accepted, or nobody is accepted, and the closer the acceptance rate $\pi$ gets to these extremes, the smaller is $d$. This is not due to a better fairness of the classifier, because the classifier is exactly the same, and its output is the same, just the classification threshold varies. We would like to assess the fairness of the classifier, therefore, similarly to the accuracy, we need to normalize the result with respect to $\pi$. 

We propose to normalize $d$ by the maximum possible $d_{\mathit{max}}$ at each $\pi$. Discrimination would be at its maximum if a classifier ranks candidates in such a way that first everyone from the favored community is accepted, and only then candidates from the protected community start to be accepted\footnote{It can be compared to a (supposedly fictional) evacuation procedure from the Titanic. Passengers are put in a queue, where all the first class passengers have a priority over third class passengers. Then as many passengers are evacuated, as there are boats.}.
In such a case the maximum discrimination is
\begin{equation}
d_{\mathit{max}} = \min \left(\frac{\pi}{\alpha},\frac{1 - \pi}{1 - \alpha} \right),
\label{eq:dmax}
\end{equation}
where $\alpha$ is the proportion of the favored community individuals in the data, and $\pi$ is the acceptance rate.

We propose to normalize the discrimination measure by the maximum possible discrimination. 
\begin{equation}
\delta = \frac{p(+|w) - p(+|b)}{d_{\mathit{max}}},
\end{equation}
where $d_{\mathit{max}}$ given in Eq.~(\ref{eq:dmax}) is the maximum possible discrimination at a given acceptance rate. 
The maximum value of $\delta$ is $1$, which means the worst possible discrimination, where the favored community has a complete priority, $\delta = 0$ means no discrimination where people from the favored and protected communities fully mix in the queue. $\delta$ can be negative, indicating a reverse discrimination.

\begin{figure}
\centering
\begin{tikzpicture}[scale=0.8]
\begin{axis}[name = plot1,
xmin= -0.05, xmax=1.05, ymin=-0.05, ymax=1.05,
legend style={draw=none,font=\footnotesize},
legend pos= north east,
height=4.3cm, width=5cm, 
xlabel = {p(+) output, $\pi$}, ylabel = {Norm. accuracy, $\kappa$}]
\addplot[mycol1,line width = 2pt]  table[x = ppred0, y = kappa0] {results_logistic.dat};
\addplot[mycol2,line width = 2pt]  table[x = ppred0, y = kr] {results_logistic.dat};
\legend{logistic r.,random};
\end{axis}
\begin{axis}[name=plot2, at=(plot1.right of south east), anchor=left of south west,
height=4.3cm, width=5cm, 
xshift=0.6cm,
xmin= -0.05, xmax=1.05, ymin=-0.05, ymax=1.05,
legend style={draw=none,font=\footnotesize},
legend pos= north east,
xlabel = {p(+) output, $\pi$}, ylabel = {Norm. discrimination, $\delta$}]
\addplot[mycol3,line width = 2pt]  table[x = ppred0, y = dov_data] {results_logistic.dat};
\addplot[mycol4,line width = 2pt]  table[x = ppred0, y = dov0] {results_logistic.dat};
\addplot[black,dashed,line width = 1pt]  table[x = ptrue, y = nn] {results_logistic.dat};
\legend{$\delta$ data, $\delta$ classifier};
\end{axis}
\end{tikzpicture} 
\caption{Normalized accuracy and discrimination.}
\label{fig:output2}
\end{figure}
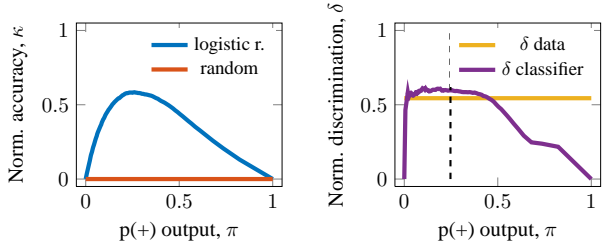

Figure \ref{fig:output2} plots normalized accuracy $\kappa$ and normalized discrimination $\delta$ of the logistic regression in our experiment. Large part of discrimination appears to be flat and closely in line with the discrimination in the data. The results now make sense, since the classifier in the experiment does not have any mechanisms for discrimination removal. At the extreme ends, where everybody is accepted, or everybody is rejected, intuitively, there is no discrimination, and the normalized measure correctly shows no discrimination.

\section{Baselines and tradeoffs}

It has been observed \cite{Kamiran10} that, assuming the labels in data are correct, discrimination removal comes at a cost -- it reduces prediction accuracy. 
The authors have found given no constraints on the acceptance rates, that the maximum possible accuracy decreases linearly with reducing difference in rates of acceptance. 
We revisit the problem of accuracy-fairness tradeoff to see if the normalized measures would show similar relations.

An oracle is a fictional baseline classifier that has the maximum possible intelligence (as if it knows the true labels), and strives to satisfy non-discrimination constraints.
A random classifier is the opposite, it does not use any intelligence. For each individual a random classifier makes a random prediction with the probability of acceptance $\pi$. 

The accuracy of the oracle will be $A_0=1$, kappa will be $\kappa_0=1$,  the discrimination would be as in the data $d_0$ and $\delta_0$. The random classifier defines the other baseline of performance with $A = \pi_0\pi + (1-\pi_0)(1-\pi)$, $\kappa = 0$, and $d = \delta = 0$. 
With $\pi = 0$ (or $\pi = 1$) the random classifier turns into the majority class classifier. 

Suppose, a decision maker aims at removing all discrimination such that $d^\star=0$ and $\delta^\star = 0$. As suggested in \cite{Kamiran10}, the oracle would either reduce the acceptance rate for the favored community (if $\alpha \leq 0.5$), or increase the acceptance rate for the protected community (if $\alpha > 0.5$). The resulting decrease in classification accuracy would be linearly proportional to the discrimination in the data $(A_0 - A) = \min \left(\alpha,(1-\alpha) \right)  (d_0 - d)$. 


We find that if the rate of acceptance is to be fixed, that is $\pi = \pi_0$, then the normalized accuracy of the oracle decreases linearly with decrease in normalized discrimination
\begin{equation}
(\kappa_0 - \kappa) = \min \left(\frac{\alpha}{\pi_0},\frac{1-\alpha}{1-\pi_0}\right)(\delta_0 - \delta).
\end{equation}

If the rate of acceptance does not need to be fixed, the optimal strategy is still the same -- either to reduce acceptance for the favored community ("decrease males"), or to increase acceptance for the protected community ("increase females"), but the choice now depends not only on $\alpha$, but also on $\pi_0$ and $\delta^\star$. We do not have a closed form solution at the moment, but Figure \ref{fig:oracle} presents simulated results of the oracle classifier on the benchmark dataset (Adult). "Change both" is the solution where the acceptance rate is kept the same as in the original data. 
These experiments show the maximum possible accuracy, given the discrimination constraints. We can see that when using the normalized measures for accuracy and discrimination the upper bounds remain linear.
\begin{figure}
\centering
\begin{tikzpicture}[scale=0.8]
\begin{axis}[name = plot1,
xmin= -0.05, xmax=0.55, ymin=0.45, ymax=1.05,
legend style={draw=none,font=\footnotesize},
legend pos= south east,
height=4.3cm, width=5cm, 
xlabel = {Discrimination, $d$}, ylabel = {Accuracy, $A$}]
\addplot[mycol1,line width = 2pt]  table[x = ddif, y = acc] {out_oracle_f.dat};
\addplot[mycol2,line width = 2pt]  table[x = ddif, y = acc] {out_oracle_m.dat};
\addplot[mycol3,line width = 2pt]  table[x = ddif, y = acc] {out_oracle_both.dat};
\legend{decrease males, increase females, change both};
\end{axis}
\begin{axis}[name=plot2, at=(plot1.right of south east), anchor=left of south west,
height=4.3cm, width=5cm, 
xshift=0.6cm,
xmin= -0.05, xmax=0.55, ymin=0.45, ymax=1.05,
legend style={draw=none,font=\tiny},
legend pos= south east,
xlabel = {Norm. discrimilnation, $\delta$}, ylabel = {Norm. accuracy, $\kappa$}]
\addplot[mycol1,line width = 2pt]  table[x = dov, y = kappa] {out_oracle_f.dat};
\addplot[mycol2,line width = 2pt]  table[x = dov, y = kappa] {out_oracle_m.dat};
\addplot[mycol3,line width = 2pt]  table[x = dov, y = kappa] {out_oracle_both.dat};
\end{axis}
\end{tikzpicture} 
\caption{Oracle.}
\label{fig:oracle}
\end{figure}


\section{Interesting cases}

We wrap up our study with an experiment to illustrate the difference between the raw and normalized measures when comparing non-discriminatory classifiers. 

The experiment compares the performance of three classifiers (logistic regression, Naive Bayes and decision tree J48 from WEKA) trained using three different strategies: including the protected characteristic among classifier inputs, excluding the protected characteristic from classifier inputs, and excluding the protected characteristic from classifier inputs plus massaging the labels of the training data. Massaging is perhaps the simplest discrimination removal strategy, it has been introduced in \cite{Kamiran09}. Training labels are converted from binary to numeric using a ranker function, we use a logistic regression fit on the same training data. A number of lowest ranked males who have a positive label are changed to negative, and the same number of highest ranked females, who have a negative label, are changed to positive such that the positive rate remains the same as in the original data, but the discrimination is zero. Then a classifier is learned on this modified training data. Testing data is not modified. Table~\ref{tab:crisp} presents the results measured on the testing data. 
\begin{table}
\caption{Performance of classifiers, everything $\times 10^{-2}$}
{\footnotesize
\begin{tabular}{llllll}
\hline
& p(+) & Acc. & Disc. & N. acc. & N. disc. \\
& $\pi$ & $A$ & $d$ & $\kappa$ & $\delta$ \\
\hline
Data/oracle & $24.7$ & $100$ & $19.9$ & $100$ & $54.4$\\
\hline
Logistic with $s$ & $20.2$ & $84.9$ & $18.3$ & $56.7$ & ${\bf 61.4}$\\
Logistic no $s$ & $20.1$ & $84.9$ & $17.6$ & $56.6$ & ${\bf 59.6}$\\
Logistic massage & $22.1$ & $83.5$ & $6.9$ & $53.9$ & $21.3$\\
\hline
NB with $s$ & $15.4$ & $81.9$ & $13.5$ & $44.2$ & ${\bf 59.7}$ \\
NB no $s$ & $14.4$ & $81.4$ & $10.9$ & $41.7$ & $51.3$ \\
NB massaged & $15.4$ & $81.5$ & $6.8$ & $43.3$ & $29.7$ \\
\hline
Tree J48 with $s$ & $19.6$ & $85.1$ & $17.9$ & $56.9$ & ${\bf 61.9}$ \\
Tree J48 no $s$ & $19.6$ & $85.0$ & $17.9$ & $56.7$ & ${\bf 61.8}$ \\
Tree massage & $22.9$ & $83.5$ & $6.1$ & $54.6$ & $18.1$ \\
\hline
\end{tabular}
}
\label{tab:crisp}
\end{table}

We can make several interesting observations. First, all classifiers tend to output lower acceptance rates than that in the original data. At the same time, if the protected characteristic is used, the discrimination measure $d$ may show a decrease in the nominal discrimination as compared to the original data, but the normalized discrimination $\delta$ by all three classifiers is even higher than in the data. Apparently, a classifier learned on discriminatory data without any protective measures amplifies discrimination. 

Removing the protected characteristic (no $s$) indicates little improvement in discrimination. This is due to, so called, redlining effect. A number of features in the data are correlated with the protected characteristic, therefore, discrimination is still captured, and, in cases of logistic regression and decision tree, is still higher than in the original dataset. 

Interestingly, massaging strategy outputs higher acceptance rates than removing the protected characteristic. The acceptance rates of massaging are closer to the positive rates in the original data, and discrimination is lower, as expected. This suggests, that when discrimination is present in the training data, but usage of the protected characteristic is not allowed, classifiers tend to decrease the acceptance rate, which may show better nominal discrimination figures, but the real underlying discrimination (measured by normalized $\delta$) remains.

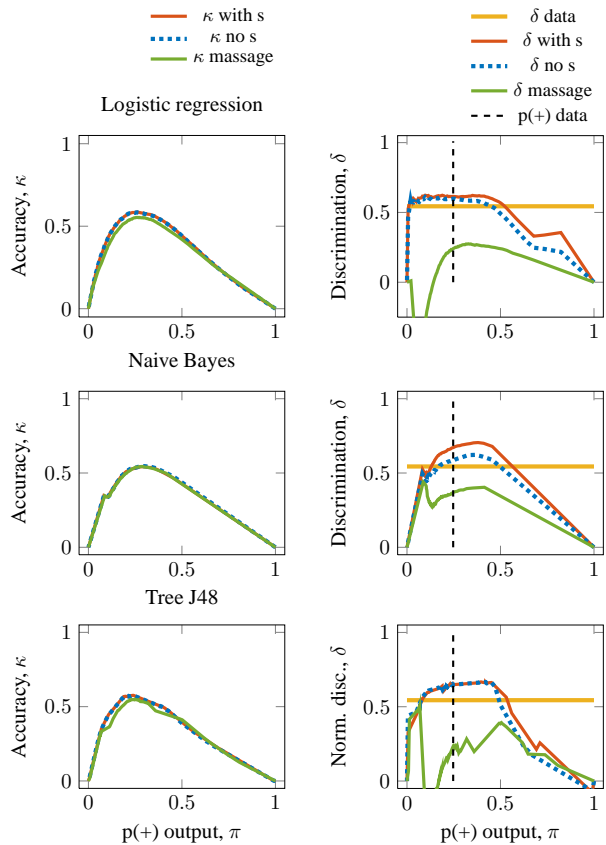
\begin{figure}
\centering
\begin{tikzpicture}[scale=0.8]
\begin{axis}[name = plot1,
xmin= -0.05, xmax=1.05, ymin=-0.05, ymax=1.05,
legend style={draw=none,font=\footnotesize,at={(1,1.75)}},
height=4.6cm, width=5cm, 
ylabel = {Accuracy, $\kappa$}, title = Logistic regression]
\addplot[mycol2,line width = 1.5pt]  table[x = ppred, y = kappa] {results_logistic.dat};
\addplot[mycol1,line width = 2pt,dotted]  table[x = ppred0, y = kappa0] {results_logistic.dat};
\addplot[mycol5,line width = 1.5pt]  table[x = ppredMS, y = kappaMS] {results_logistic.dat};
\legend{$\kappa$ with s, $\kappa$ no s, $\kappa$ massage};
\end{axis}
\begin{axis}[name=plot2, at=(plot1.right of south east), anchor=left of south west,
height=4.6cm, width=5cm, 
xshift=0.6cm,
xmin= -0.05, xmax=1.05, ymin=-0.25, ymax=1.05,
legend style={draw=none,font=\footnotesize,at={(1,1.75)}}, 
ylabel = {Discrimination, $\delta$}]
\addplot[mycol3,line width = 2pt]  table[x = ppred, y = dov_data] {results_logistic.dat};
\addplot[mycol2,line width = 1.5pt]  table[x = ppred, y = dov] {results_logistic.dat};
\addplot[mycol1,line width = 2pt,dotted]  table[x = ppred0, y = dov0] {results_logistic.dat};
\addplot[mycol5,line width = 1.5pt]  table[x = ppredMS, y = dovMS] {results_logistic.dat};
\addplot[black,dashed,line width = 1pt]  table[x = ptrue, y = nn] {results_logistic.dat};
\legend{$\delta$ data, $\delta$ with s,$\delta$ no s, $\delta$ massage, p(+) data}; 
\end{axis}
\begin{axis}[name=plot3, at=(plot1.below south west), anchor=above north west,
height=4.3cm, width=5cm, 
xmin= -0.05, xmax=1.05, ymin=-0.05, ymax=1.05,
ylabel = {Accuracy, $\kappa$}, title = Naive Bayes]
\addplot[mycol2,line width = 1.5pt]  table[x = ppred, y = kappa] {results_SNB.dat};
\addplot[mycol1,line width = 2pt,dotted]  table[x = ppred0, y = kappa0] {results_SNB.dat};
\addplot[mycol5,line width = 1.5pt]  table[x = ppredMS, y = kappaMS] {results_SNB.dat};
\end{axis}
\begin{axis}[name=plot4, at=(plot3.right of south east), anchor=left of south west,
height=4.3cm, width=5cm, 
xshift=0.6cm,
xmin= -0.05, xmax=1.05, ymin=-0.05, ymax=1.05,
ylabel = {Discrimination, $\delta$}]
\addplot[mycol3,line width = 2pt]  table[x = ppred, y = dov_data] {results_SNB.dat};
\addplot[mycol2,line width = 1.5pt]  table[x = ppred, y = dov] {results_SNB.dat};
\addplot[mycol1,line width = 2pt,dotted]  table[x = ppred0, y = dov0] {results_SNB.dat};
\addplot[mycol5,line width = 1.5pt]  table[x = ppredMS, y = dovMS] {results_SNB.dat};
\addplot[black,dashed,line width = 1pt]  table[x = ptrue, y = nn] {results_SNB.dat};
\end{axis}
\begin{axis}[name=plot5, at=(plot3.below south west), anchor=above north west,
height=4.3cm, width=5cm, 
xmin= -0.05, xmax=1.05, ymin=-0.05, ymax=1.05,
legend style={draw=none,font=\tiny},
legend pos= north east,
xlabel = {p(+) output, $\pi$}, ylabel = {Accuracy, $\kappa$}, title = Tree J48]
\addplot[mycol2,line width = 1.5pt]  table[x = ppred, y = kappa] {results_J48.dat};
\addplot[mycol1,line width = 2pt,dotted]  table[x = ppred0, y = kappa0] {results_J48.dat};
\addplot[mycol5,line width = 1.5pt]  table[x = ppredMS, y = kappaMS] {results_J48.dat};
\end{axis}
\begin{axis}[name=plot6, at=(plot5.right of south east), anchor=left of south west,
height=4.3cm, width=5cm, 
xshift=0.6cm,
xmin= -0.05, xmax=1.05, ymin=-0.05, ymax=1.05,
xlabel = {p(+) output, $\pi$}, ylabel = {Norm. disc., $\delta$}]
\addplot[mycol3,line width = 2pt]  table[x = ppred, y = dov_data] {results_J48.dat};
\addplot[mycol2,line width = 1.5pt]  table[x = ppred, y = dov] {results_J48.dat};
\addplot[mycol1,line width = 2pt,dotted]  table[x = ppred0, y = dov0] {results_J48.dat};
\addplot[mycol5,line width = 1.5pt]  table[x = ppredMS, y = dovMS] {results_J48.dat};
\addplot[black,dashed,line width = 1pt]  table[x = ptrue, y = nn] {results_J48.dat};
\end{axis}
\end{tikzpicture} 
\caption{Performance of baseline classifiers.}
\label{fig:compare}
\end{figure}

Finally, Figure \ref{fig:compare} presents normalized accuracies and discriminations at different acceptance rates. Overall we can see that massaging does remove some of discrimination, but at many acceptance rates the removal is not very precise, and sometimes even overshoots introducing a reverse discrimination. This calls for a revision of the massaging, and possibly other discrimination removal techniques, taking into consideration possibility of different acceptance rates and normalized measures of discrimination.
%

\section{Conclusion}

Evaluation of non-discriminatory classifiers needs to take into account positive output rates, otherwise the comparison may be misleading and conclusions about comparative performance may be invalid. 

We have introduced a normalization factor for discrimination measure, considering the maximum possible discrimination at a given acceptance rate. The maximum discrimination is present when the protected individuals start to be accepted only after everybody from the favored community is accepted.

Acceptance rates may be constrained by resources, and not freely available to choose for decision makes. If the acceptance rate in the data and in the classifier outputs is fixed, then classifiers are comparable in terms of $A$ and $d$, otherwise they need to be compared in terms of $\kappa$ and $\delta$.

%

\newpage 
\bibliography{bib_discrimination}
\bibliographystyle{icml2015}

\end{document}